CHAPTER 45

# ROBOTS AS POWERFUL ALLIES FOR THE STUDY OF EMBODIED COGNITION FROM THE BOTTOM UP

MATEJ HOFFMANN AND ROLF PFEIFER

## Introduction

The study of human cognition—and human intelligence—has a long history and has kept scientists from various disciplines—philosophy, psychology, linguistics, neuroscience, artificial intelligence, and robotics—busy for many years. While there is no agreement on its definition, there is wide consensus that it is a highly complex subject matter that will require, depending on the particular position or stance, a multiplicity of methods for its investigation. Whereas, for example, psychology and neuroscience favor empirical studies on humans, artificial intelligence has proposed computational approaches, viewing cognition as information processing, as algorithms over representations. Over the last few decades, overwhelming evidence has been accumulated showing that the pure computational view is severely limited and that it must be extended to incorporate embodiment, i.e., the agent's somatic setup and its interaction with the real world, and, because they are real physical systems, robots became the tools of choice to study cognition. There have been a plethora of pertinent studies, but they all have their own intrinsic limitations. In this chapter, we demonstrate that a robotic approach, combined with information theory and a developmental perspective, promises insights into the nature of cognition that would be hard to obtain otherwise.

We start by introducing "low-level" behaviors that function without control in the traditional sense; we then move to sensorimotor processes that incorporate reflex-based loops (involving neural processing). We discuss "minimal cognition" and show how the role of embodiment can be quantified using information theory, and we introduce the so-called SMCs, or sensorimotor contingencies, which can be viewed as the very basic building blocks of cognition. Finally, we expand on how humanoid robots can be productively exploited to make inroads in the study of human cognition.

## Behavior Through Interaction

What cognitive scientists are regularly forgetting is that complex coordinated behaviors—for example, walking, running over uneven terrain, swimming, avoiding obstacles—can often be realized with no or minimal involvement of cognition/representation/computation. This is possible because of the properties of the body and the interaction with the environment, that is, the embodied and embedded nature of the agent. Robotics is well suited for providing existence proofs of this kind and then to further analyze these phenomena. We will only briefly present some of the most notable case studies.

### Low-Level Behavior: Mechanical Feedback Loops

A classical illustration of behavior in complete absence of a "brain" is the passive dynamic walker (McGeer 1990): a minimal robot that can walk without any sensors, motors, or control electronics. It loosely resembles a human, with two legs, no torso, and two arms attached to the "hips," but its ability to walk is exclusively due to the downward slope of the incline on which it walks and the mechanical parameters of the walker (mainly leg segment lengths, mass distribution, foot shape, and frictional characteristics). The walking movement is entirely the result of finely tuned mechanics on the right kind of surface. A motivation for this research is also to show how human walking is possible with minimal energy use and only limited central control. However, most of the problems that animals or robots are faced with in the real world cannot be solved solely by passive interaction of the physical body with the environment. Typically, active involvement by means of muscles/motors is required. Furthermore, the actuation pattern needs to be specified by the agent,[1] and hence a controller of some sort is required. Nevertheless, it turns out that if the physical interaction of the body with the environment is exploited, the control program can be very simple. For example, the passive dynamic walker can be modified by adding a couple of actuators and sensors and a reflex-based controller, resulting in the expansion of its niche to level ground while keeping the control effort and energy expenditure to a minimum (Collins et al. 2005).

However, in the real world, the ground is often not level and frequent corrective action needs to be taken. It turns out that often the very same mechanical system can generate this corrective response. This phenomenon is known as *self-stabilization* and is a result of a mechanical feedback loop. To use dynamical systems terminology, certain trajectories (such as walking with a particular gait) have attracting properties and small perturbations are automatically corrected, without control—or one could say that "control" is inherent in the mechanical system.[2] Blickhan et al. (2007) review self-stabilizing properties of biological muscles in a paper entitled "Intelligence by Mechanics"; Koditschek et al. (2004) analyze walking insects and derive inspiration for the design of a hexapod robot with unprecedented mobility (RHex—e.g., Saranli et al. 2001).

[1] In this chapter, we will use "agent" to describe humans, animals, or robots.

## Sensorimotor Intelligence

Mechanical feedback loops constitute the most basic illustration of the contribution of embodiment and embeddedness to behavior. The immediate next level can be probably attributed to direct, reflex-like, sensorimotor loops. Again, robots can serve to study the mechanisms of "reactive" intelligence. Grey Walter (Walter 1953), the pioneer of this approach, built electronic machines with a minimal "brain" that displayed phototactic-like behavior. This was picked up by Valentino Braitenberg (Braitenberg 1986) who designed a whole series of two-wheeled vehicles of increasing complexity. Even the most primitive ones, in which sensors are directly connected to motors (exciting or inhibiting them), display sophisticated behaviors. Although the driving mechanisms are simple and entirely deterministic, the interaction with the real world, which brings in noise, gives rise to complex behavioral patterns that are hard to predict.

This line was picked up by Rodney Brooks, who added an explicit anti-representationalist perspective in response to the in-the-meantime-firmly-established cognitivistic paradigm (e.g., Fodor 1975; Pylyshyn 1984) and "good old-fashioned artificial intelligence" (GOFAI) (Haugeland 1985). Brooks openly attacked the GOFAI position in the seminal articles "Intelligence without Reason" (Brooks 1991a) and "Intelligence without Representation" (Brooks 1991b), and proposed *behavior-based robotics* instead. Through building robots that interact with the real world, such as insect robots (Brooks 1989), he realized that "when we examine very simple level intelligence we find that explicit representations and models of the world simply get in the way. It turns out to be better to use the world as its own model" (Brooks 1991b). Inspired by biological evolution, Brooks created a decentralized control architecture consisting of different layers; every layer is a more or less simple coupling of sensors to motors. The levels operate in parallel but are built in a hierarchy (hence the term *subsumption architecture*; Brooks 1986). The individual modules in the architecture may have internal states (the agents are thus not purely reactive any more); however, Brooks argued against calling the internal states representations (Brooks 1991b).

[2] The description is idealized—in reality, a walking machine would fall into the category of "hybrid dynamical systems," where the notions of attractivity and stability are more complicated.

# Minimal Embodied Cognition

In the case studies described in the previous section, the agents were either mere physical machines or they relied on simple direct sensorimotor loops only—resembling reflex arcs of the biological realm. They were reactive agents constrained to the "here-and-now" time scale, with no capacity for learning from experience and also no possibility of predicting the future course of events. Although remarkable behaviors were sometimes demonstrated, there are intrinsic limitations.

The introduction of first instances of internal simulation, which goes beyond the "here-and-now" time scale, is considered the hallmark of cognition by some (e.g., Clark and Grush 1999). This could be a simple forward model (as present already in insects—see Webb 2004) that provides the prediction of a future sensory state given the current state and a motor command (efference copy). Forward models could provide a possible explanation of the evolutionary origin of first simulation/emulation circuitry[3] and of environmentally decoupled thought—the agent employing primitive "models" before or instead of directly operating on the world.

> Early emulating agents would then constitute the most minimal case of what Dennett calls a Popperian creature—a creature capable of some degree of off-line reasoning and hence able (in Karl Popper's memorable phrase) to "let its hypotheses die in its stead" (Dennett 1995, p. 375). (Clark and Grush 1999, p. 7)

Importantly, we are still far from any abstract models or symbolic reasoning. Instead, we are dealing with the sensorimotor space and the possibility for the agent to extract regularities in it and later exploit this experience in accordance with its goals. For example, the agent can learn that given a certain visual stimulation, say, from a cup, a particular motor action (reach and grasp) will lead to a pattern of sensory stimulation (in humans: we can feel the cup in the hand). The sensorimotor space plays a key part here and it is critically shaped by the embodiment of the agent and its embedding in the environment: a specific motor signal only leads to a distinct result if embedded in the proper physical setup. If you change the shape and muscles of the arm, the motor signal will not result in a successful grasp.

## Quantifying the Effect of Embodiment Using Information Theory

For cognitive development of an agent, the "quality" of the sensorimotor space determines what can be learned. First, the type of sensory receptors—their mechanism of transduction—determines what kind of signals the agent's brain or controller will be receiving from the environment. Furthermore, the shape and placement of these sensors will perform an additional transformation of the information that is available in the environment.

[3] See Grush (2004) for the similarities and differences between emulation theory (Grush 2004) and simulation theory (Jeannerod 2001).

For example, different species of insects have evolved different non-homogeneous arrangements of the light-sensitive cells in their eyes, providing an advantageous non-linear transformation of the input for a particular task. One example is exploiting ego-motion together with motion parallax to gauge distance to objects in the environment and eventually facilitate obstacle avoidance. Using a robot modeled after the facet eye of a housefly, Franceschini et al. (1992) showed that the nonlinear arrangement of the facets—more dense in the front than on the side—compensates for the motion parallax and allows uniform motion detection circuitry to be used in the entire eye, which makes it easy for the robot to avoid obstacles with little computation. These findings were confirmed in experiments with artificial evolution on real robots (Lichtensteiger 2004). Artificial eyes with designs inspired by arthropods include Song et al. (2013) and Floreano et al. (2013).

It is not always possible to pinpoint the specific transformation of sensory signals that is facilitated by the morphology as in the previous case. A more general tool is provided by the methods of information theory. Information is used in the Shannon sense here—to quantify statistical patterns in observed variables. The structure or amount of information induced by particular sensor morphology could be captured by different measures, for example, entropy. However, information (structure) in the sensory variables tells only half of the story (a "passive perception" one in this case), because organisms interact with their environments in a closed-loop fashion: sensory inputs are transformed into motor outputs, which in turn determine what is sensed next. Therefore, the "raw material" for cognition is constituted by the sensorimotor variables and it is thus crucial to study relationships between sensors and motors, as illustrated by the sensorimotor contingencies (see next section). Furthermore, time is no less important a variable. Lungarella and Sporns (2006) provide an excellent example of the use of information theoretic measures in this context. In a series of experiments with a movable camera system, they could show that, for example, the entropy in the visual field is decreased if the camera is tracking a moving visual target (a red ball) compared to the condition where the movement of the ball and the camera were uncorrelated. This is intuitively plausible, because if the object is kept in the center of the visual field, there is more "order," i.e., less entropy. A collection of case studies on information-theoretic implications of embodiment in locomotion, grasping, and visual perception is presented by Hoffmann and Pfeifer (2011).

## Sensorimotor Contingencies

Sensorimotor contingencies (SMCs) were originally presented in the influential article by O'Regan and Noë (2001) as the structure of the rules governing sensory changes produced by various motor actions. The SMCs, according to O'Regan and Noë, are the key "raw material" upon which perception, cognition, and eventually consciousness operates. Furthermore, they sketch a possible hierarchy ranging from modality-related (or apparatus-related) SMCs to object-related SMCs. The former, the modality-related SMCs, would capture the immediate effect that certain actions (or movements) have on sensory stimulation. Clearly, these would be sensory modality specific (e.g., head movement will induce a different change in the SMCs of the visual and auditory modalities—turning the head will change the visual stimulation almost entirely, whereas changes in the acoustic system will be minimal) and would strongly depend on the sensory morphology. Therefore, this concept is strongly related to what we have discussed in the previous sections: (1) different sensory morphology importantly affects the information flow induced in the sensory receptors and hence also the corresponding SMCs; (2) the effect of action is already constitutively included in the SMC notion itself.

Although conceptually very powerful, the notion of SMCs was not articulated concretely enough in O'Regan and Noë (2001) such that it could be expressed mathematically or directly transferred into a robot implementation, for example. Bührmann et al. (2013) have proposed a formal dynamical systems account of SMCs. They devised a dynamical system description for the environment and the agent, which is in turn split into body, internal state (such as neural activity), motor, and sensory dynamics. Bührmann et al. are making a distinction between sensorimotor (SM) environment, SM habitat, SM coordination, and SM strategy. The SM environment is the relation between motor actions and changes in sensory states, independent of the agent's internal (neural) dynamics. The other notions—from SM habitat to SM strategies—add internal dynamics to the picture. SM habitat refers to trajectories in the sensorimotor space, but subject to constraints given by the internal dynamics that are responsible for generating motor commands, which may depend on previous sensory states as well—an example of closed-loop control. SM coordination then further reduces the set of possible SM trajectories to those "that contribute functionally to a task." For example, specific patterns of squeezing an object in order to assess its hardness would be SM coordination patterns serving object discrimination. Finally, SM strategies take, in addition, "reward" or "value" for the agent into account.

As wonderfully illustrated by Beer and Williams (2015), the dynamical systems and information theory are two complementary mathematical lenses through which brain–body–environment systems can be studied. While acknowledging the merits of both frameworks as "intuition, theory, and experimental pumps" (Beer and Williams 2015), it is probably fair to say that compared to dynamical systems, information theory has been thus far more successfully applied to the analysis of real systems of higher dimensionality. This is true for both natural systems—in particular, brains (Garofalo et al. 2009; Quiroga and Panzeri 2009)—and artificial systems. Thus, to study sensorimotor contingencies in a real robot beyond the simple simulated agents of Bührmann et al. (2013) and Beer and Williams (2015), we chose to use the lens of information theory. Following up on related studies of e.g., Olsson et al. (2004), we conducted a series of studies in a real quadrupedal robot with rich nonlinear dynamics and a collection of sensors from different modalities (Hoffmann et al. 2012; Hoffmann et al. 2014; Schmidt et al. 2013) (see Box 45.1). We have applied the notion of "transfer entropy"

## Box 45.1 Sensorimotor contingencies in a quadruped robot

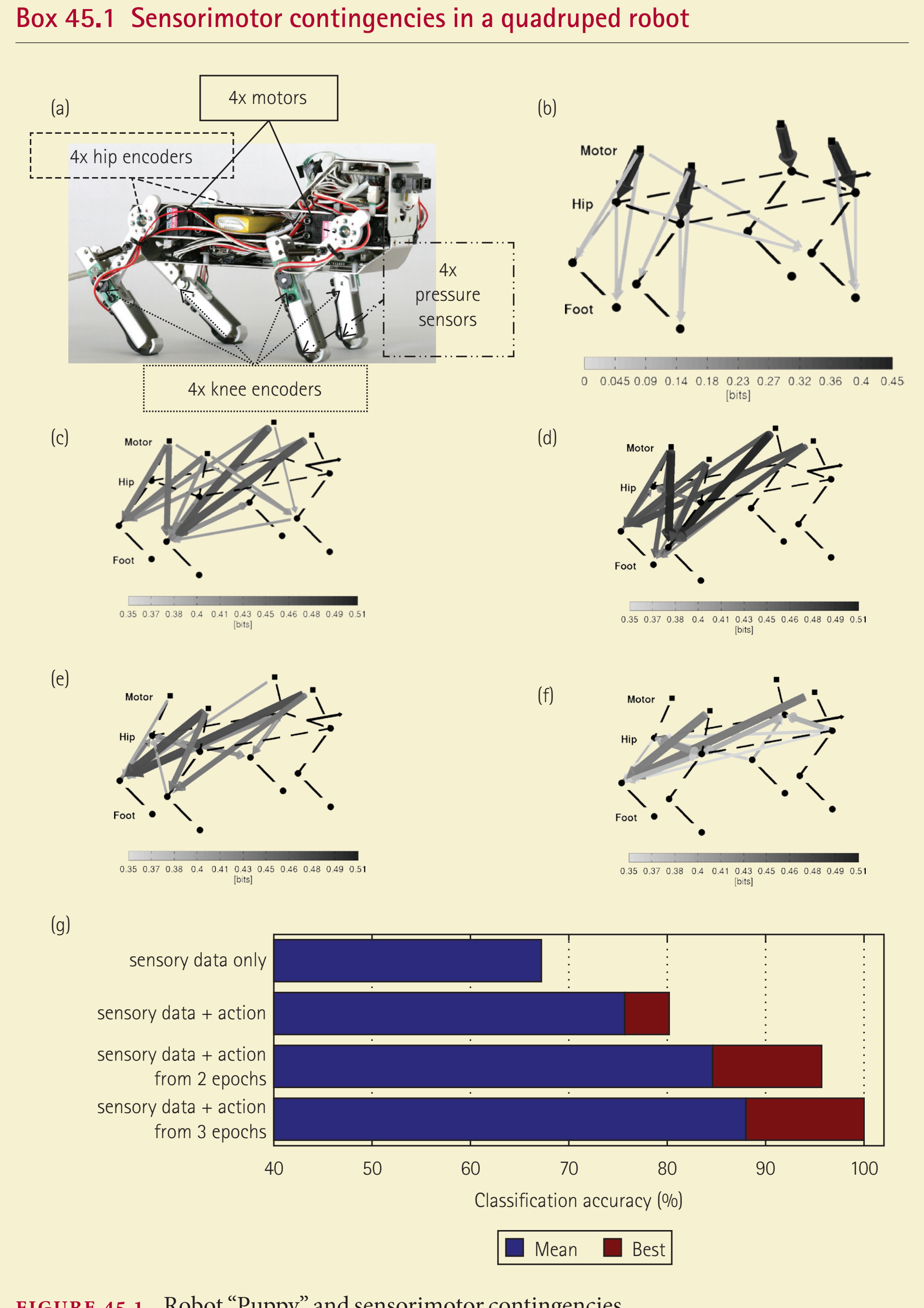

**FIGURE 45.1.** Robot "Puppy" and sensorimotor contingencies.

Experiments were conducted on the quadrupedal robot Puppy (Figure 45.1a), which has four servomotors in the hips together with encoders measuring the angle at the joint, four encoders in the passive compliant knees, and four pressure sensors on the feet. We used the notion of "transfer entropy" from information theory, which can be used to measure directed information flows between time series. In our case, the time series were collected from individual motor and sensory channels and the information transfer was calculated for every pair of channels two times, once in every direction (say, from hind right motor to front right knee encoder and also in the opposite direction). Loosely speaking, transfer entropy from channel A to channel B measures how well the future state of channel B can be predicted knowing the current state of channel A (see Schmidt et al. 2013 for details).

First, we wanted to investigate the "sensorimotor structure," i.e., the relative strengths of relationships between different sensors and motors, which is intrinsic to the robot's embodiment (body + sensor morphology only). To this end, random motor commands were applied and the relationships between motor and sensory variables were studied, closely resembling the notion of SM environment (Bührmann et al. 2013). The strongest information flows between pairs of channels were extracted and are shown overlaid over the schematic of the Puppy robot (dashed lines) in panel B. The transfer entropy is encoded as thickness and gray level of the arrows. The strongest flow occurs from the motor signals to their respective hip joint angles, which is clear because the motors directly drive the respective hip joints. The motors have a smaller influence on the knee angles (stronger in the hind legs) and on the feet pressure sensors—on the respective legs where the motor is mounted, thus illustrating that body topology was successfully extracted (at the same time, the flows from the hind leg motors and hips to the front knees highlight that the functional relationships are different than the static body structure; see also Schatz and Oudeyer 2009). These patterns are analogous to the modality-related SMCs; just as we can predict what will be the sensory changes induced by moving the head, the robot can predict the effects of moving the hind leg, say.

In a second step, we studied the relationships in the sensorimotor space when the robot was running with specific coordinated periodic movement patterns or gaits. The results for two selected gaits—turn left and bound right*—are shown in panels C and D, respectively. The flows from motors to the hip joints, which would again dominate, were left out of the visualization. The plots clearly demonstrate the important effect of specific action patterns in two ways. First, they markedly differ from the random motor command situation: the dominant flows are different and, in addition, the magnitude of the information flows is bigger (the number of bits—note the different range of the color bar compared to B), illustrating how much information structure is induced by the "neural pattern generator." Second, they also significantly differ between themselves. The "turn left" gait in panel C reveals the dominant action of the right leg and in particular the knee joint. In the "bound right" gait in D, the motor signals are predictive of the sensory stimulation in the hind knees and also the left foot. The gaits were obtained by optimizing the robot's performance for speed or for turning and thus correspond to patterns that are functionally relevant for the robot and can even be said to carry "value." Thus, in the perspective of Bührmann et al. (2013), our findings about the sensorimotor space using the gaits can be interpreted as studying the SM coordination or even SM strategy of the quadruped robot.

Finally, next to the embodiment or morphology (shape of the body and limbs, type and placement of sensors and effectors, etc.) and the brain (the neural dynamics responsible for generating the coordinated motor command sequences), the SMCs are co-determined by the environment as well. All the results thus far came from sensorimotor data collected from the robot running on a plastic foil ground (low friction). Panels E and F depict how the information flows for the bound right gait are modulated when the robot runs on a different ground (E—Styrofoam, F—rubber). The overall pattern is similar to D, but the flows to the left foot disappear, and eventually flows to the left knee joint become dominant. This is because the posture of the robot changed: the left foot contacts the ground at a different angle now, inducing less stimulation in the pressure sensor. Also, as the friction increases (from the foil over Styrofoam to rubber), the push-off during stance of the left hind leg becomes stronger, resulting in more pronounced bending of the knee. Finally, since the high-friction ground poses more resistance to the robot's movements, the trajectories are less smooth and the overall information flow drops.

While all the components (body, brain, environment) have a profound effect on the overall sensorimotor space, our analysis reveals that in this case, the gait used (as prescribed primarily by the "neural/brain" dynamics) is a more important factor than the environment (the ground)—the latter seems to modulate the basic structure of information flows induced by the gait. This has important consequences for the agent when it is to learn something about its environment and perform perceptual categorization, for example. In order to investigate this quantitatively, we have presented the robot with a terrain (the surface/ground it was running on) classification task. Relying on sensory information alone leads to significantly worse terrain classification results than when the gait is explicitly taken into account in the classification process (Hoffmann, Stepanova, and Reinstein 2014). Furthermore, in line with the predictions of the sensorimotor contingency theory, longer sensorimotor sequences are necessary for object perception (Maye and Engel 2012). That is, while in short sequences (motor command, sensory consequence), modality-related SMCs (panel B) will be dominant, longer interactions will allow objects the agent is interacting with to stand out. Using data from our robot, this is convincingly demonstrated in panel G. The first row shows classification results when using data from one sensory epoch (two seconds of locomotion) collapsed across all gaits, i.e., without the action context. Subsequent rows report results where classification was performed separately for each gait and increasingly longer interaction histories were available. "Mean" values represent the mean performance; "best" are classification results from the gait that facilitated perception the most (see Hoffmann et al. 2012 for details).

* "Turn left" was a movement pattern dominated by the action of the right hind leg that was pushing the robot forward and left. Regarding "bound right," bounding gait is a running gait used by small mammals. It is similar to gallop, and features a flight phase, but is characterized by synchronous action of every pair of legs. However, in this study, we used lower speeds without an aerial phase. In addition, the symmetry of the motor signals was slightly disrupted, resulting in a right-turning motion.

from information theory, which can be used to characterize sensorimotor flows in the robot—for example, how strongly sensors are affected by motor commands—and we tried to isolate the effects of the body, motor programs (gaits), and environment in the agent's sensorimotor space. Finally, we tested the predictions of SMC theory regarding object discrimination. In our investigations, we have chosen the situated perspective—analyzing only the relationships between sensory and motor variables that would also be available to the agent itself. However, information-theoretic methods can also be productively applied to study relationships between internal and external variables, such as between sensory or neuronal states and some properties of an external object (e.g., its size, Beer and Williams 2015; or any other property that can be expressed numerically). Using this approach, one can obtain important insights into the operation and temporal evolution of categorization, for example. Performing this in the ground discrimination scenario on the quadrupedal robot constitutes our future work.

While the studies on "minimally cognitive agents" are of fundamental importance and lead to valuable insights for our understanding of intelligent behavior, the ultimate target is, of course, human cognition. Toward this end, one may want to resort to more sophisticated tools, for example, humanoid robots.

## Human-like Cognition in Robots

In the previous section, we showed how robots can be beneficial in operationalizing, formalizing, and quantifying ideas, concepts, and theories that are important for understanding cognition but that are often not articulated in sufficient detail. An obvious implication of this analysis is that the kind of cognition that emerges will be highly dependent on the body of the agent, its sensorimotor apparatus, and the environment it is interacting with. Thus, to target human cognition, the robot's morphology—shape, type of sensors, and their distribution, materials, actuators—should resemble that of humans as closely as possible. Now we have to be realistic: approximating humans very closely would imply mimicking their physiology, the sensors in the body, and the inner organs, the muscles with comparable biological instantiation, and the bloodstream that supplies the body with energy and oxygen. Only then could the robot experience the true concept, e.g., of being thirsty or out of breath, hearing the heart pumping, blushing, or the feeling of quenching the thirst while drinking a cold beer in the summer. So, even if, on the surface, a robot might be almost indistinguishable from a human (like, for example, Hiroshi Ishiguro's recent humanoid "Erica"), we have to be aware of the fundamental differences: comparatively very few muscles and tendons, no actuators that can get sore when overused, no sensors for pain, only low-density haptic sensors, no sweat glands in the skin, and so on and so forth. Thus, "Erica" will have a very impoverished concept of drinking or feeling hot. In other words, we have to make substantial abstractions.

Just as an aside, making abstractions is nothing bad—in fact, it is one of the most crucial ingredients of any scientific explanation because it forces us to focus on the essentials, ignoring whatever is considered irrelevant (the latter most likely being the majority of things that we could potentially take into account). Thus, the specifics of the robot's cognition—its concepts, its body schema—will clearly diverge from that of humans, but the underlying principles will, at a certain level of abstraction, be the same. For example, it will have its own sensorimotor contingencies, it will form cross-modal associations through Hebbian learning, and it will explore its environment using its sensorimotor setup. So if the robot says "glass," this will relate to very different specific sensorimotor experiences, but if the robot can recognize, fill, and hand a "glass" to a human for drinking, it makes sense to say that the robot has acquired the concept of "glass."

Because the acquisition of concepts is based on sensorimotor contingencies, which in turn require actions on the part of the agent, and because the patterns of sensory stimulation are associated with the respective motor signals, the robot platforms of choice will ideally be tendon-driven—just like humans who use muscles and tendons for movements. Given our discussion on abstraction earlier, we can also study concept acquisition in robots that have motors in the joints—we just have to be aware of the concrete differences. Still, the principles governing the robot's cognition can be very similar to that of humans (see Box 45.2 for examples of different types of humanoid robots).

## BOX 45.2 Humanoid embodiment for modeling cognition

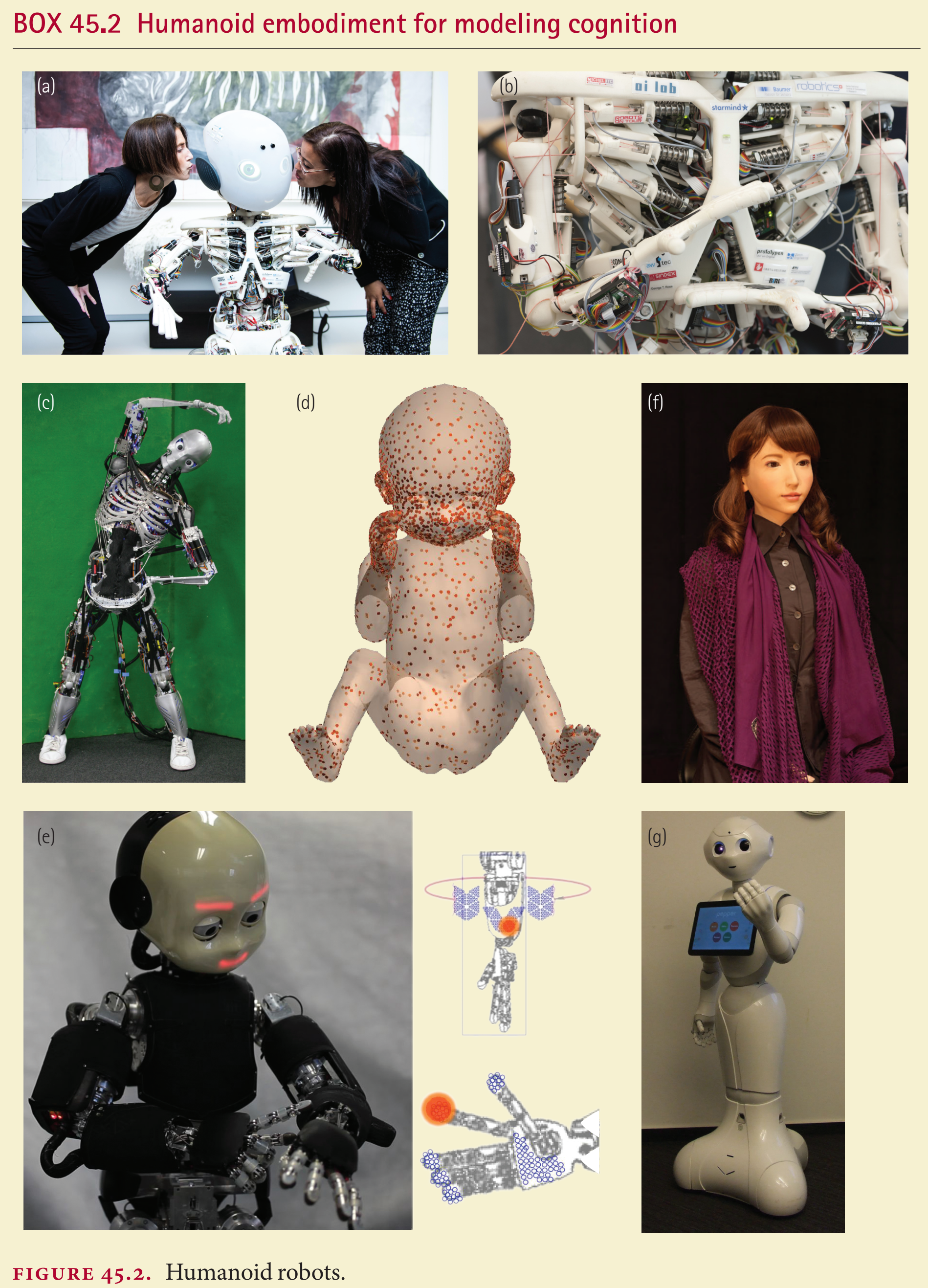

**FIGURE 45.2.** Humanoid robots.

A large number of humanoid robots have been developed over the last decades and many of them can, one way or other, be used to study human cognition. Given that all of them to date are very different from real humans—each of them, implicitly or explicitly, embodies certain types of abstractions—there is no universal platform, but they have all been developed with specific goals in mind. Here we present a few examples and discuss the ways in which they are employed in trying to ferret out the principles of human cognition. The categories shown in Figure 45.2 are musculoskeletal robots (Roboy and Kenshiro), "baby" robots with sensorized skins (iCub and fetus simulators), and social interaction robots (Erica and Pepper).

In order to use the robots for learning their own complex dynamics and for building up a body schema, both Roboy and Kenshiro (Nakanishi et al. 2012) need to be equipped with many sensors so that they can "experience" the effect of a particular actuation pattern. Given rich sensory feedback, using the principle that every action leads to sensory stimulation, both these robots can, in principle, employ motor babbling in order to learn how to move. Especially for Kenshiro, with his very large number of muscles, learning is a must. A very important step in this direction is the work of Richter et al. (2016), who have combined a musculoskeletal robotics toolkit (Myorobotics) with a scalable neuromorphic computing platform (SpiNNaker) and demonstrated control of a musculoskeletal joint with a simulated cerebellum.

Finally, if the interest is social interaction, it might be more productive to use robots like Erica or Pepper. Both Erica and Pepper are somewhat limited in their sensorimotor abilities (especially haptics), but are endowed with speech understanding and generation facilities; they can recognize faces and emotions; and they can realistically display any kind of facial expression.

**Musculoskeletal robots: Roboy and Kenshiro**

Figure 45.2a. Roboy overview: The musculoskeletal design can be clearly observed. At this point, Roboy has 48 "muscles." Eight are dedicated to each of the shoulder joints. This can no longer be sensibly programmed: learning is a necessity. Currently, Roboy serves as a research platform for the EU/FET Human Brain Project to study, among other things, the effect of brain lesions on the musculoskeletal system. Because it has the ability to express a vast spectrum of emotions, it can also be employed to investigate human–robot interaction, and as an entertainment platform.

Credit: © Embassy of Switzerland in the United States of America.

Figure 45.2b. Close-up of the muscle-tendon system. Although the shoulder joint is distinctly dissimilar to a human one—for example, it doesn't have a shoulder blade—it is controlled by eight muscles, which require substantial skills in order to move properly: which muscles have to be actuated to what extent in order to achieve a desired movement?

Credit: © Erik Tham/Corbis Documentary/Getty Images.

Figure 45.2c. Kenshiro's musculoskeletal setup. The musculoskeletal design is clearly visible. At this point, Kenshiro has 160 "muscles"—50 in the legs, 76 in the trunk, 12 in the shoulder, and 22 in the neck. In terms of musculoskeletal system, it is the one robot that most closely resembles the human. So, if learning of the dynamics in this system is the goal, Kenshiro will be the robot of choice. Note that although Kenshiro is "closest" to a human in this respect, it is still subject to enormous abstractions. Currently, Kenshiro serves as a research platform at the University of Tokyo to investigate tendon-controlled systems with very many degrees of freedom (Nakanishi et al. 2012).

Credit: Photo courtesy Yuki Asano.

**"Baby" robots with sensitive skins**

Figure 45.2d. Fetus simulator. A musculoskeletal model of human fetus at 32 weeks of gestation has been constructed and coupled with a brain model comprising 2.6 million spiking neurons (Yamada et al. 2016). The figure shows the tactile sensor distribution, which was based on human two-point discrimination data.

Reproduced from Yasunori Yamada, Hoshinori Kanazawa, Sho Iwasaki, Yuki Tsukahara, Osuke Iwata, Shigehito Yamada, and Yasuo Kuniyoshi, An Embodied Brain Model of the Human Foetus, *Scientific Reports*, 6 (27893), Figure 1d, doi:10.1038/srep27893

Figure 45.2e. The iCub baby humanoid robot. The iCub (Metta et al. 2010) has the size of a roughly four-year-old child and corresponding sensorimotor capacities: 53 degrees of freedom (electrical motors), two stereo cameras in a biomimetic arrangement, and over 4,000 tactile sensors covering its body. The panel shows the robot performing self-touch and corresponding activations in the tactile arrays of the left forearm and right index finger.

**Social interaction robots: Erica and Pepper**

Figure 45.2f. Erica, the latest creation of Prof. Hiroshi Ishiguro, was designed specifically with the goal of imitating human speech and body language patterns, in order to have "highly natural" conversations. It also serves as a tool to study human–robot interaction, and social interaction in general. Moreover, because of its close resemblance to humans, the "uncanny valley"—the fact that people get uneasy when the robots are too human-like—hypothesis can be further explored and analyzed (see, e.g., Rosenthal-von der Pütten, Marieke, and Weiss 2014, where the Geminoid HI-1 modeled after Prof. Ishiguro was used).

Credit: Photo courtesy of Hiroshi Ishiguro Laboratory, ATR and Osaka University.

Figure 45.2g. Pepper, a robot developed by Aldebaran (now Softbank Robotics), although much simpler (and much cheaper!) than Erica, is used successfully on the one hand to study social interaction, for entertainment, and to perform certain tasks (such as selling Nespresso machines to customers in Japan).

## The Role of Development

A very powerful approach to deepen our understanding of cognition, and one that has been around for a long time in psychology and neuroscience, is to study ontogenetic development. During the past two decades or so, this idea has been adopted by the robotics community and has led to a thriving research field dubbed "developmental robotics." Now, a crucial part of ontogenesis takes place in the uterus. There, tactile sense is the first to develop (Bernhardt 1987) and may thus play a key role in the organism's learning about first sensorimotor contingencies, in particular, those pertaining to its own body (e.g., hand-to-mouth behaviors). Motivated by this fact, Mori and Kuniyoshi (2010) developed a musculoskeletal fetal simulator with over 1,500 tactile receptors, and studied the effect of their distribution on the emergence of sensorimotor behaviors. Importantly, with a natural (non-homogeneous) distribution, the fetus developed "normal" kicking and jerking movements (i.e., similar to those observed in a human fetus), whereas with a homogeneous allocation it did not develop any of these behaviors. Yamada et al. (2016), using a similar fetal simulator and a large spiking neural network brain model, have further studied the effects of intrauterine (vs. extrauterine) sensorimotor experiences on cortical learning of body representations. A physical version—the fetusoid—is currently under development (Mori et al. 2015). Somatosensory (tactile and proprioceptive) inputs continue to be of key importance also in early infancy when "infants engage in exploration of their own body as it moves and acts in the environment. They babble and touch their own body, attracted and actively involved in investigating the rich intermodal redundancies, temporal contingencies, and spatial congruence of self-perception" (Rochat 1998, p. 102). The iCub baby humanoid robot (Metta et al. 2010) (Box 45.2E), equipped with a whole-body tactile array (Maiolino et al. 2013) comprising over 4,000 elements, is an ideal platform to study these processes. The study of Roncone et al. (2014) on self-calibration using self-touch is a first step in this direction.

## Applications Of Human-Like Robots

Finally, this research strand—employing humanoid robots to study human cognition—has also important applications. In traditional domains and conventional tasks—such as pick-and-place operations in an industrial environment—current factory automation robots are doing just fine. However, robots are starting to leave these constrained domains, entering environments that are far less structured and are starting to share their living space with humans. As a consequence, they need to dynamically adapt to unpredictable interactions and guarantee their own as well as others' safety at every moment. In such cases, more human-like characteristics—both physical and "mental"—are desirable. Box 45.3 illustrates how more brain-like body representations can help robots to become more autonomous, robust, and safe. The possibilities for future applications of robots with cognitive capacities are enormous, especially in the rapidly

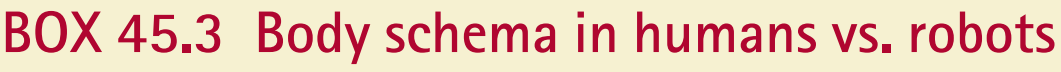

## BOX 45.3 Body schema in humans vs. robots

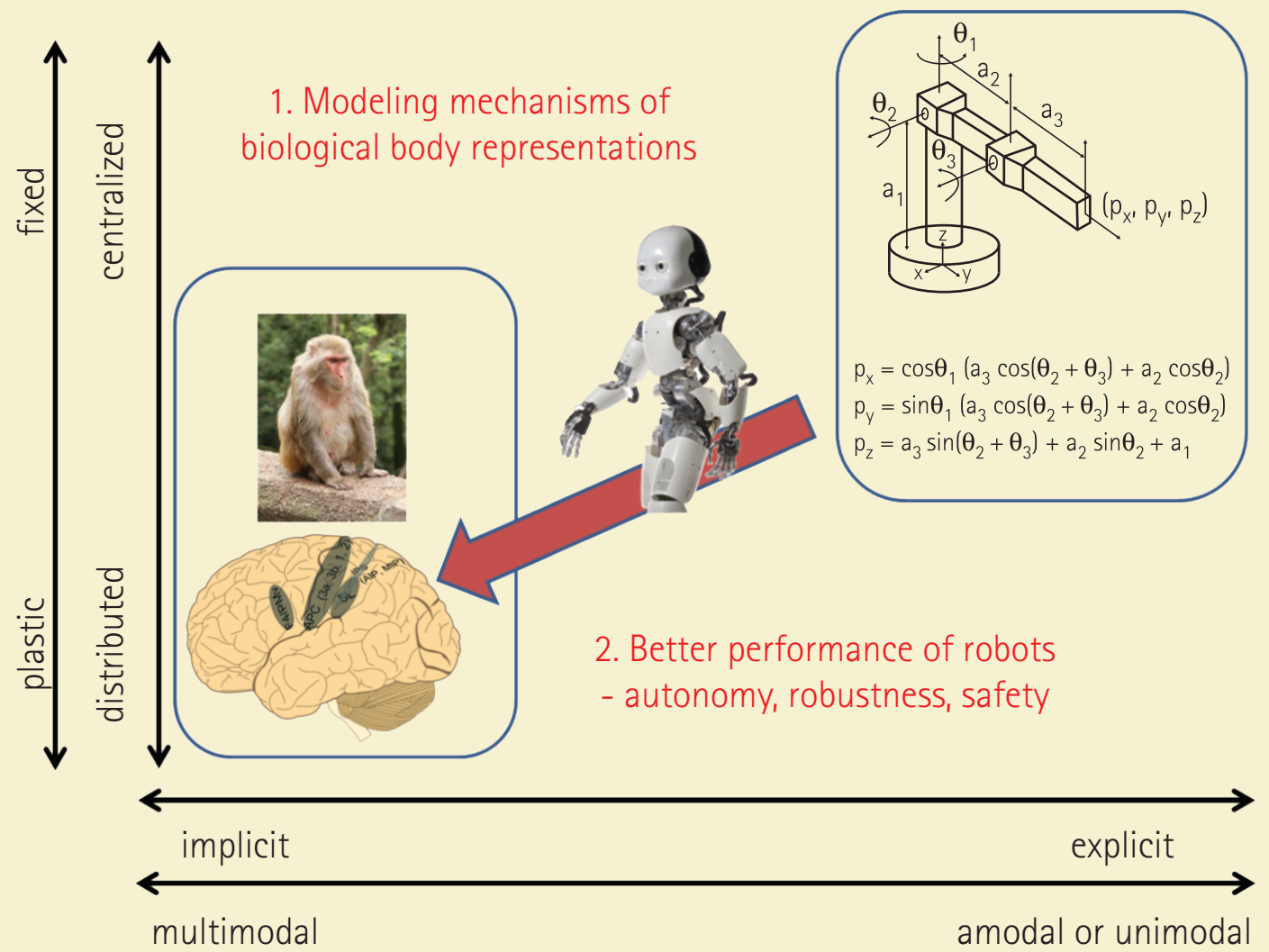

**FIGURE 45.3.** Characteristics of body representations.

Credit: Monkey photo source: Einar Fredriksen/Flickr/Attribution-ShareAlike 4.0 International (CC BY-SA 4.0)
Credit: Brain image source: Hugh Guiney/Attribution-ShareAlike 3.0 Unported (CC BY-SA 3.0)
Credit: Line drawing and equations source: Reproduced with the permission of Dr. Hugh Jack from http://www.engineeronadisk.com
Credit: iCub Robot source: © iCub Facility—IIT, 2017

A typical example of a traditional robot and its mathematical model is depicted in the upper right of Figure 45.3. The robot is an arm consisting of three segments with three joints between the base and the final part—the end-effector. Its model is below the robot—the forward kinematics equations that relate configuration of the robot (joint positions $\theta_1$, $\theta_2$, $\theta_3$) to the Cartesian position of the end-effector ($p_x$, $p_y$, $p_z$). The model has the following characteristics: (1) it is explicit—there is a one-to-one correspondence between its body and the model ($a_1$ in the model is the length of the first arm segment, for example); (2) it is unimodal—the equations directly describe physical reality; one sensory modality (proprioception—joint angle values) is needed to get the correct mapping in the current robot state; (3) it is centralized—there is only one model that describes the whole robot; (4) it is fixed—normally, this mapping is set and does not change during the robot operation. Other models/mappings are typically needed for robot operation, such as inverse kinematics, differential kinematics, or models of dynamics (dealing with forces and torques), but they would all share the above-mentioned characteristics (see Hoffmann et al. 2010 for a survey).

As pointed out earlier, animals and humans have different bodies than robots; they also have very different ways of representing them in their brains. The panel in the lower left shows the rhesus macaque and below some of the key areas of its brain that deal with body representations (see, e.g., Graziano and Botvinick 2002). There is ample evidence that these representations differ widely from the ones traditionally used in robotics—namely, "the body in the brain" would be (1) implicitly represented—there would hardly be a "place" or a "circuit" encoding, say, the length of a forearm; such information is most likely only indirectly available and possibly in relation to other variables; (2) multimodal—drawing mainly from somatosensory (tactile and proprioceptive) and visual, but also vestibular (inertial) and closely coupled to motor information; (3) distributed—there are numerous distinct, but partially overlapping and interacting representations that are dynamically recruited depending on context and task; (4) plastic—adapting over both long (ontogenesis) and short time scales, as adaptation to tool use (e.g., Iriki et al. 1996) or various body illusions testify (e.g., humans start feeling ownership over a rubber hand after minutes of synchronous tactile stimulations of the hand replica and their real hand under a table; Botvinick and Cohen 1998).

The iCub robot "walking" from the top right to the bottom left in the figure is illustrating two things. First, in order to be able to model the mechanisms of biological body representations, the traditional robotic models are of little use—a radically different approach needs to be taken. Second, by making the robot models more brain-like, we hope to inherit some of the desirable properties typical of how humans and animals master their highly complex bodies. Autonomy and robustness or resilience are one such case. It is not realistic to think that conditions, including the body, will stay constant over time and a model given to the robot by the manufacturer will always work. Inaccuracies will creep in due to wear and tear and possibly even more dramatic changes can occur (e.g., a joint becomes blocked). Humans and animals display a remarkable capacity for dealing with such changes: their models dynamically adapt to muscle fatigue, for example, or temporarily incorporate objects like tools after working with them, or reallocate "brain territory" to different body parts in case of amputation of a limb. Robots thus also need to perform continuous self-modeling (Bongard et al. 2006) in order to cope with such changes. Finally, unlike factory robots that blindly execute their trajectories and thus need to operate in cages, humans and animals use multimodal information to extend the representation of their bodies to the space immediately surrounding them (also called peripersonal space). They construct a "margin of safety," a virtual "bubble" around their bodies that allows them to respond to potential threats such as looming objects, warranting safety for them and also their surroundings (e.g., Graziano and Cooke 2006). This is highly desirable in robots as well, and can transform them from dangerous machines to collaborators possessing whole-body awareness like we do. First steps along these lines in the iCub were presented by Roncone et al. (2016).

growing area of service robotics, where robots perform tasks in human environments. Rather than accomplishing them autonomously, they often do it in cooperation with humans, which constitutes a big trend in the field. In cooperative tasks, it is of course crucial that the robots understand the common goals and the intentions of the humans in order to be successful. In other words, they require substantial cognitive skills. We have barely started exploiting the vast potential of these types of cognitive machines.

## Conclusion

Our analysis so far has demonstrated that robots fit squarely into the embodied and pragmatic (action-oriented) turn in cognitive sciences (e.g., Engel et al. 2013), which implies that whole behaving systems rather than passive subjects in brain scanners need to be studied. Robots provide the necessary grounding to computational models of the brain by incorporating the indispensable brain–body–environment coupling (Pezzulo et al. 2011). The advantage of synthetic methodology, or "understanding by building" (Pfeifer and Bongard 2007), is that one learns a lot in the process of building the robot and instantiating the behavior of interest. The theory one wants to test thus automatically becomes explicit, detailed, and complete. Robots become virtual experimental laboratories retaining all the virtues of "theories expressed as simulations" (Cangelosi and Parisi 2002), but bring the additional advantage that there is no "reality gap": there is real physics and real sensory stimulation, which lends more credibility to the analysis if embodiment is at center stage.

We are convinced that robots are the right tools to help us understand the embodied, embedded, and extended nature of cognition because their makeup—physical artifacts with sensors and actuators interacting with their environment—automatically warrants the necessary ingredients. It seems that they are particularly suited for investigations of cognition from bottom up (Pfeifer et al. 2014), where development under particular constraints in brain–body–environment coupling is crucial (e.g., Thelen and Smith 1994). It also becomes possible to simulate conditions that one would not be able to test in humans or animals—think of the simulation of fetal ontogenesis while manipulating the distribution of tactile receptors (Mori and Kuniyoshi 2010). Furthermore, many additional variables (such as internal states of the robot) become easily accessible and lend themselves to quantitative analysis, such as using methods from information theory. Therefore, the combination of a robot with sensorimotor capacities akin to humans, the possibility of emulating the robot's growth and development, and finally the ease of access to all internal variables that can be subject to rigorous quantitative investigations create a very powerful tool to help us understand cognition.

We want to close with some thoughts on whether it is possible to realize—next to embodied, embedded, and extended—enactive robots as well. Most researchers in embodied AI/cognitive robotics automatically adopt the perspective of extended functionalism (Clark 2008; Wheeler 2011), whereby the boundaries of cognitive systems can be extended beyond the agent's brain and even skin—including the body and environment. However, it has been pointed out by the proponents of enactive cognitive science (Di Paolo 2010; Froese and Ziemke 2009) that in order to fully understand cognition in its entirety, embedding the agent in a closed-loop sensorimotor interaction with the environment is necessary, yet may not be sufficient in order to induce important properties of biological agents such as intentional agency. In other words, one should not only study instances of individual closed sensorimotor loops as models of biological agents—that would be the recommendation of Webb (2009)—but one should also try to endow the models (robots in this case) with similar properties and constraints that biological organisms are facing. In particular, it has been argued that life and cognition are tightly interconnected (Maturana 1980; Thompson 2007), and a particular organization of living systems—which can be characterized by autopoiesis (Maturana 1980) or metabolism, for example—is crucial for the agent to truly acquire meaning in its interactions with the world. While these requirements are very hard to satisfy with the artificial systems of today, Di Paolo (2010) proposes a way out: robots need not metabolize, but they should be subject to so-called precarious conditions. That is, the success of a particular instantiation of sensorimotor loops or neural vehicles in the agent is to be measured against some viability criterion that is intrinsic to the organization of the agent (e.g., loss of battery charge, overheating leading to electronic board problems resulting in loss of mobility, etc.). The control structure may develop over time, but the viability constraint needs to be satisfied, otherwise the agent "dies" (McFarland and Boesser 1993). In a similar vein, in order to move from embodied to enactive AI, Froese and Ziemke (2009) propose to extend the design principles for autonomous agents of Pfeifer and Scheier (2001), requiring the agents to generate their own systemic identity and regulate their sensorimotor interaction with the environment in relation to a viability constraint. The unfortunate implication, however, is that research along these lines will in the short term most likely not produce useful artifacts. On the other hand, this approach may eventually give rise to truly autonomous robots with unimaginable application potential.

## Acknowledgments

M.H. was supported by a Marie Curie Intra European Fellowship (iCub Body Schema 625727) within the 7th European Community Framework Programme and the Czech Science Foundation under Project GA17-15697Y.